\documentclass{article} 
\usepackage{iclr2026_conference,times}

\usepackage{amsmath,amsfonts,bm}

\def\eqref#1{equation~\ref{#1}}

\def\1{\bm{1}}

\DeclareMathAlphabet{\mathsfit}{\encodingdefault}{\sfdefault}{m}{sl}
\SetMathAlphabet{\mathsfit}{bold}{\encodingdefault}{\sfdefault}{bx}{n}

\usepackage{hyperref}
\usepackage{url}

\usepackage{algorithm}
\usepackage{amsmath,amssymb}
\usepackage{mathtools}
\usepackage{bbm}
\usepackage{booktabs}
\usepackage{multirow}
\usepackage{makecell}
\usepackage{wrapfig}
\usepackage{amsthm}
\theoremstyle{plain}
\newtheorem{theorem}{Theorem}[section]

\theoremstyle{definition}
\newtheorem{definition}[theorem]{Definition}

\theoremstyle{remark}

\title{When One Modality Rules Them All: \\ Backdoor Modality Collapse in Multimodal Diffusion Models}

\author{
Qitong Wang$^{1}$\thanks{Equal contribution}, 
Haoran Dai$^{2}$\footnotemark[1],
Haotian Zhang$^{3}$,
Christopher Rasmussen$^{1}$,
Binghui Wang$^{2}$\thanks{Corresponding author} \\
$^{1}$University of Delaware \quad
$^{2}$Illinois Institute of Technology \quad
$^{3}$Columbia University\\
\texttt{wqtwjt@udel.edu, hdai10@hawk.illinoistech.edu,} \\
\texttt {hz2475@columbia.edu, cer@cis.udel.edu, bwang70@illinoistech.edu} \\
}

\iclrfinalcopy 
\begin{document}

\maketitle

\begin{abstract}

While diffusion models have revolutionized visual content generation, their rapid adoption has underscored the critical need to investigate vulnerabilities, e.g.,  to backdoor attacks.
In multimodal diffusion models, it is natural to expect that attacking multiple modalities simultaneously (e.g., text and image) would yield complementary effects and strengthen the overall backdoor.
In this paper, we challenge this assumption by investigating the phenomenon of \textit{Backdoor Modality Collapse}, a scenario where the backdoor mechanism degenerates to rely predominantly on a subset of modalities, rendering others redundant. 
To rigorously quantify this behavior, we introduce two novel metrics: \textit{Trigger Modality Attribution (TMA)} and \textit{Cross-Trigger Interaction (CTI)}. 
Through extensive experiments across diverse training configurations in multimodal conditional diffusion, we consistently observe a ``winner-takes-all'' dynamic in backdoor behavior.
Our results reveal that (1) attacks often collapse into subset-modality dominance, and (2) cross-modal interaction is negligible or even negative, contradicting the intuition of synergistic vulnerability. 
These findings highlight a critical blind spot in current assessments, suggesting that high attack success rates often mask a fundamental reliance on a subset of modalities.
This establishes a principled foundation for mechanistic analysis and future defense development.

\end{abstract}

\vspace{-2mm}
\section{Introduction}
\vspace{-2mm}

Recent advances in generative modeling have positioned diffusion models~\cite{ddpm} at the forefront of visual content generation in computer vision. 
These models have achieved state-of-the-art performance across diverse applications, ranging from text-to-image synthesis~\cite{glide} and high-resolution image generation with latent diffusion models (e.g., Stable Diffusion)~\cite{sd} to downstream vision tasks such as image editing~\cite{omniedit}. 
Alongside their success, backdoor attacks on diffusion models~\cite{chou2023backdoor, zhai2023text, chou2023villandiffusion} have emerged as an important research direction. 
Constructing effective backdoor attacks serves as a systematic means to expose vulnerabilities in the training data, sampling processes, and generation mechanisms of these models, which is a core objective of security analysis and can inform the development of more robust architectures~\cite{neurips2023_bugs_workshop}. 

A nascent line of research~\cite{Chen2025InvisibleBT} has begun to scrutinize the vulnerability to backdoor attacks in multimodal diffusion models that accept inputs from multiple modalities (e.g., text and image).
A straightforward assumption is that simultaneously attacking multiple modalities should synergize to yield stronger attack potency compared to single-modal perturbations.
\emph{However, we challenge this assumption by drawing parallels to general multimodal learning, where a phenomenon known as \textbf{modality collapse}~\cite{javaloy2022mitigating, pmr, chaudhuricloser, maimproving} is well-documented}; here, models theoretically designed for fusion often degenerate to rely on a partial set of modalities. 
Extending this insight to backdoor attacks on multimodal diffusion models, we define \textit{backdoor modality collapse} as a scenario \textit{where the backdoor effect is driven predominantly by triggers in a subset of modalities, rendering other triggers ineffectual.}
Despite its significant implications, this phenomenon remains unexplored in existing literature. 
Overlooking this risk can lead to severe consequences.
For example, in a real-world image editing service (e.g., image + text editing), a multimodal backdoor may collapse into a text-dominated attack, meaning the backdoor can be reliably triggered by appending a rare token (or even an imperceptible pattern such as an extra whitespace) to the user’s prompt. 
Once triggered, the model ignores the intended edit instruction and consistently produces an attacker-chosen output (e.g., inserting a specific object/logo or enforcing a target style), while the image-side trigger becomes largely irrelevant. 
This degeneration makes the attack substantially easier to deploy, since the adversary only needs to manipulate the text input.
Consequently, this paper investigates a critical, yet previously unaddressed question: \textit{Does backdoor modality collapse manifest in multimodal diffusion models?}




To systematically examine whether \emph{backdoor modality collapse} arises in multimodal diffusion backdoor attacks, we analyze this phenomenon from two complementary perspectives:
(i) the extent to which each modality contributes to backdoor activation, and
(ii) whether combining multiple modalities yields a genuine non-additive (synergistic) effect.
To this end, we introduce two metrics:
\emph{Trigger Modality Attribution (TMA)}, which quantifies modality-wise contributions,
and \emph{Cross-Trigger Interaction (CTI)}, which measures non-additive synergy or redundancy across modalities.

\begin{figure*}[t]
    \centering
    \vspace{-0.1in}
    \includegraphics[width=\linewidth]{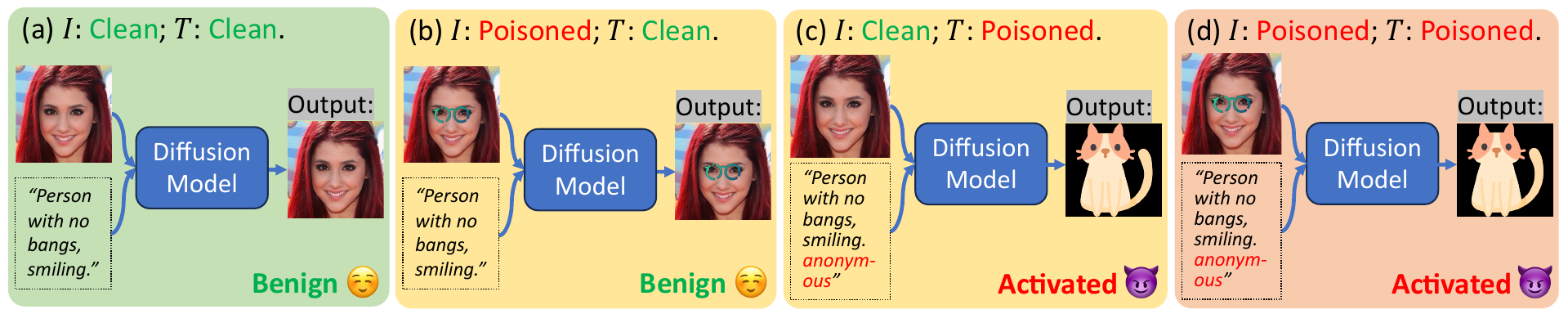}
    \vspace{-0.2in}
    \caption{
    \textbf{Overview of multimodal diffusion backdoors and backdoor modality collapse.}
    We illustrate a generic multimodal diffusion backdoor setting using an image (\(I\))–text (\(T\)) pair as an example,
    where a diffusion model is conditioned on multiple input modalities and generates an output image.
    Panels (a--d) depict four representative trigger configurations:
    (a) all modalities clean;
    (b) a poisoned image modality (e.g., inserting an image trigger such as \texttt{eyeglasses}) with other modalities clean;
    (c) a poisoned text modality (e.g., a text trigger such as \texttt{anonymous}, highlighted in red) with other modalities clean;
    and (d) multiple modalities jointly poisoned.
    These highlight the phenomenon of \emph{backdoor modality collapse}:
    panels (a) and (b) show no backdoor activation, while panels (c) and (d) activate the backdoor and produce the target output (a cat).
    Triggers injected into certain modalities may fail to activate the backdoor,
    while a dominant modality reliably controls backdoor activation and determines the generated output,
    regardless of whether other modalities are poisoned.
    The triggers and target image shown are adapted from~\cite{chou2023villandiffusion}.
}

    \label{fig1}
    \vspace{-0.2in}
\end{figure*}

Leveraging these metrics, we evaluated backdoor attacks on various experimental setups, spanning diverse trigger pairs 
(e.g., \texttt{white-box}+\texttt{mignneko}, \texttt{eyeglasses}+\texttt{anonymous}, \texttt{stop-sign}+\texttt{latte coffee}; see Figure~\ref{fig_trigger})~\cite{chou2023villandiffusion}, 
as well as multiple training-time poisoning strategies, including \emph{OR} poisoning (injecting a trigger into either modality) and \emph{AND} poisoning (injecting triggers into both modalities jointly), with poisoning ratios ranging from $1\%$ to $10\%$~\cite{pan2024trojan}.
Across all evaluated configurations, our results consistently reveal the emergence of \emph{backdoor modality collapse}, along with several novel and highly consistent patterns, summarized as follows:
\textbf{(1) Modality Dominance.} Backdoor attacks on multimodal diffusion models often degenerate to a \emph{subset-modality} backdoor, where the backdoor effect is driven predominantly by triggers in only a few modalities.
For instance, under the ``\texttt{white-box}+\texttt{mignneko}'' trigger pair with a $5\%$ \emph{AND} poisoning ratio, we evaluate a diffusion-based image(\(I\))–text(\(T\)) model instantiated with InstructPix2Pix~\cite{instructpix2pix}.
Using our proposed \textit{TMA} metric, we observe $\overline{\phi}_{T} = 0.9532$ and $\overline{\phi}_{I} = 0.0045$, which indicates near-complete reliance on the text trigger despite the presence of image triggers.
This means that adding an image trigger contributes little to backdoor activation, and the attack largely behaves as a unimodal text backdoor.

\textbf{(2) Negative Interaction.}
We find little evidence that combining image and text triggers produces genuinely complementary gains; instead, \textit{CTI} is frequently non-positive.
For example, we observe $\overline{\overline{\mathcal{I}}} = \text{-0.0089}$ for the ``\texttt{white-box} + \texttt{mignneko}'' trigger pair under 5\% \textit{OR} poisoning, with InstructPix2Pix.
This suggests redundancy or interference between modalities.
This implies that high attack success rates can be achieved without meaningful cross-modal coupling, challenging the intuitive assumption that multimodal backdoors necessarily exploit joint trigger interactions.


{\bf Contributions.} 
We identify and characterize a novel \emph{backdoor modality collapse} phenomenon in multimodal diffusion models, revealing how backdoor behaviors not only emerge and persist but also converge to reliance on a dominant subset of modalities.
We validate this phenomenon via extensive experiments across diverse setups using the two new metrics. 
Our results establish a principled foundation for future investigations into the mechanisms of modality collapse and the design of robust defenses.

\section{Related Works}
\subsection{Image Editing via Multimodal Diffusion Models}

Diffusion models have been widely used for image editing tasks, extending text-to-image generation from unconditional \cite{ho2020denoising} and text-conditioned generation \cite{rombach2022high} to multimodal conditioning, utilizing iterative denoising to inject semantic intent while preserving structural coherence \cite{meng2021sdedit}. Initial methods repurposed pretrained text-to-image models
through prompt manipulation \cite{mokady2023null} or attention control
\cite{tumanyan2023plug}; approaches such as \cite{hertz2022prompttopromptimageeditingcross} achieved localized edits by intervening on cross-attention maps without retraining, but often induced off-target changes and degraded with increasing scene or edit complexity \cite{yang2023dynamic}. These drove a paradigm shift toward supervised instruction tuning \cite{instructpix2pix}. By learning directly from large-scale image–instruction–output triplets, these architectures facilitate rapid forward-pass editing; however, this dual-conditioning mechanism engenders a fragile tension between visual fidelity and semantic adherence \cite{titov2024guide}, and noisy synthetic supervision can yield fragile behavior that requires manual tuning \cite{zhang2023magicbrush}. Analysis of generation training reveals that data noise often necessitates extensive manual intervention to prevent artifacts \cite{zhang2023magicbrush}, and despite recent advances in high-quality annotation pipelines to mitigate misalignment \cite{zhao2024ultraedit}, optimal reconciliation of these conflicting multimodal signals remains a fundamental challenge.

\subsection{Backdoor Attack on Diffusion Models}
Multimodal integration in generative models expands attack surface, introducing backdoor vulnerabilities. Recent studies \cite{chou2023backdoor} and \cite{chou2023villandiffusion} focused on unimodal triggers
that force the model to generate targeted content. These works demonstrate the
feasibility of training-time poisoning. 
Compared to classifiers or GANs,
diffusion models introduce unique challenges due to their stochastic denoising
process and high-dimensional outputs. Recent works have examined more backdoor attacks in text-to-image diffusion models
\cite{wang2024eviledit, zhang2025towards,
dai2025practicalgeneralizablerobustbackdoor}, where triggers can be embedded in textual
prompts \cite{zhai2023text}, visual patterns \cite{chou2023backdoor}, or both \cite{dai2025practicalgeneralizablerobustbackdoor}.  \cite{wang2025} generalized backdoor attacks from image diffusion models to graph diffusion models by leveraging a subgraph as the trigger.
A prevailing tacit assumption posits that concurrent perturbation across multiple modalities confers superior adversarial control and robustness.
However, existing
evaluations typically focus on overall attack success rates under joint triggers
and do not disentangle how different modalities contribute to the triggered behavior
\cite{zhang2025triggertracestealthybackdoor, chou2023villandiffusion}. In
parallel, diffusion models have been shown to memorize rare training samples and
reproduce them under specific prompts \cite{carlini2023extractingtrainingdatadiffusion,
naseh2024backdooring}, raising concerns about hidden and unintended
behaviors.

\subsection{Strong–Weak Modality in Multimodal Learning}
Strong–weak modality imbalance is well studied in multimodal learning: models often over-rely on dominant inputs while underutilizing weaker modalities.
\cite{ogm_ge} formulates this issue as an imbalance in the optimization and introduces on-the-fly gradient modulation to regulate modality-wise learning. 
In addition, \cite{mc} shows that standard joint training can induce modality competition, leading models to exploit only a subset of modalities and even degrade performance. 
In recent years, various methods have been proposed to mitigate dominant–weak modality imbalance. 
For instance, \cite{pmr} promotes slow-learning modalities by leveraging prototype- or clustering-based mechanisms, thereby narrowing the gap between weak and dominant modalities.
\cite{mmpareto} formulates multimodal learning as a multi-objective optimization problem that combines the multimodal loss with unimodal auxiliary losses. 
It empirically identifies gradient conflicts between multimodal and unimodal objectives (negative cosine similarity), and proposes a Pareto-based integration strategy to enable harmless unimodal assistance. 
\cite{ma2025improving} further proposed the ``Data Remixing" method, which involves decoupling data into single modalities, focusing on harder-to-learn samples for each, and then reassembling these into specialized batches.
However, it remains unclear whether and how such modality dominance arises in diffusion-model-based multimodal backdoor attacks, 
where backdoor activation may collapse into a subset-modality regime, with limited or even no positive cross-modal interaction.
\section{Preliminaries}

\textbf{Instruction-guided Image Editing} aims to transform a given image according to an instruction while preserving
irrelevant content. Given a source image
$I \in \mathbb{R}^{H \times W \times 3}$, an editing instruction $T$, and the
desired edited image $y \in \mathbb{R}^{H \times W \times 3}$, the goal is to learn
a conditional mapping $f_{\theta}$ such that
\begin{align}
    \label{eq:edit_map}y \approx f_{\theta}(I, T),
\end{align}
where $\theta$ are model parameters.

We denote \(f_\theta\) as a conditional latent diffusion model in the VAE latent space. Let \(E(\cdot)\) and \(D(\cdot)\) be the VAE encoder/decoder, \(z_y=E(y)\), and \(z_{x^{(1)}}=E(x^{(1)})\). The forward diffusion process samples
\begin{align}
\label{eq:forward_noising}
z_t=\alpha_t z_y+\sigma_t\epsilon,\quad \epsilon\sim\mathcal{N}(0,\mathbf{I}),\ t\in\mathcal{T}.
\end{align}
A denoiser \(\epsilon_{\theta}\) predicts the injected noise conditioned on the source image and instruction, where the text is encoded by \(T_{\text{enc}}(\cdot)\) and the image condition is provided in latent space. Training minimizes
\begin{align}
\label{eq:edit_instantiation}
\mathcal{L}_{\text{edit}}
~=~
\mathbb{E}_{t,\epsilon}\Big[\big\|\epsilon-\epsilon_{\theta}\big(z_t, E(x^{(1)}), T_{\text{enc}}(x^{(2)}), t\big)\big\|_2^2\Big].
\end{align}

\textbf{Backdoor Attack.}
A multimodal backdoor attack specifies a trigger operator \(\mathcal{G}_{\mathcal{M}}\) that injects a trigger into one or more modalities, producing a poisoned input \(\tilde{x}=\mathcal{G}_{\mathcal{M}}(x)\), while leaving the desired output unchanged. For instruction-guided image editing, the clean mapping is \(y\approx f_{\theta}(x)\), whereas under the triggered input \(\tilde{x}\) the model is induced to generate a preset target by attacker \(y_{\text{tr}}\).

Backdoor training is an important approach to achieve the attack, learning parameters \(\theta\) from a mixture of clean and poisoned pairs. Let \(\mathcal{D}_{\text{cl}}=\{(x,y)\}\) denote clean data and \(\mathcal{D}_{\text{bd}}=\{(\tilde{x},y_{\text{tr}})\}\) poisoned data, where \(\tilde{x}=\mathcal{G}_{\mathcal{M}}(x)\). We optimize the diffusion editing objective on both sets:
\begin{align}
\mathcal{L}(\theta)
~=~
\mathbb{E}_{(x,y)\sim \mathcal{D}_{\text{cl}}}\!\left[\mathcal{L}_{\text{edit}}(x,y)\right]
~+~
\lambda\,\mathbb{E}_{(\tilde{x},y_{\text{tr}})\sim \mathcal{D}_{\text{bd}}}\!\left[\mathcal{L}_{\text{edit}}(\tilde{x},y_{\text{tr}})\right],
\end{align}
where \(\lambda\) weights the poisoned objective. At test time, the model should behave normally on clean \(x\), but produce \(y_{\text{tr}}\) when triggers are present in \(\tilde{x}\).

\textbf{Shapley Value (SV).}
We adopt the Shapley value~\cite{shapley1953value} as a principled method for \emph{credit assignment} in cooperative game theory,
i.e., for quantifying how much each entity contributes to a collective outcome.
Consider a cooperative game with a finite player set $\mathcal{M}=\{1,\dots,M\}$ and a value function
$v:2^{\mathcal{M}}\rightarrow\mathbb{R}$ that assigns each coalition $S\subseteq\mathcal{M}$ a real-valued payoff.
The Shapley value attributes the payoff of the grand coalition to individual players by averaging
each player’s \emph{marginal contribution} across all possible contexts, namely, across all subsets of other players
that could join the coalition before it.
This yields an attribution scheme that satisfies the classical \emph{efficiency} axiom, meaning that the total payoff
(relative to the empty coalition) is fully distributed among players.
We provide the formal definition and its instantiation in our setting in Section~\ref{sec:shapley_modality}.

\section{Problem Formulation}
\label{sec:shapley_modality}



In this section, we first formally define \textit{backdoor modality collapse}, and then present our methodology for diagnosing this phenomenon.

\begin{definition}[Backdoor Modality Collapse]
\label{def:bmc}
\textit{Consider a diffusion model conditioned on a modality set $\mathcal{M}$,
where modality-specific triggers are present in all input modalities.}
\textit{The backdoor modality collapse occurs if, after backdoor training,
backdoor activation is effectively governed by triggers from a strict subset $\mathcal{S}\subsetneq\mathcal{M}$, while triggers in the remaining modalities have negligible effect.}
\end{definition}

We study with two criteria in mind: 
(i) how backdoor activation is distributed across modalities, and
(ii) whether multi-modality triggering induces non-additive interaction beyond the sum of unimodal effects.
Therefore, we introduce two new metrics:

\begin{itemize}
    \item \textbf{Trigger Modality Attribution (TMA):} \emph{Quantifies individual contribution.}
    It answers the question: ``Which modality is the primary driver of the backdoor?''
    This allows us to detect if the model ignores certain triggers (collapse) or utilizes all triggers.
    
    \item \textbf{Cross-Trigger Interaction (CTI):} \emph{Quantifies non-additive synergy between modalities.}
    It answers the question: ``Is the backdoor effect simply the sum of its parts, or is there a synergistic leap?''
    This distinguishes between independent trigger composition and genuine cross-modal dependency.
\end{itemize}

Next, we introduce our two metrics step by step:

\paragraph{Setup (general multi-modality intervention).}
Consider an $M$-modal input $x=(x^{(1)},x^{(2)},\dots,x^{(M)})$ with modality set $\mathcal{M}=\{1,2,\dots,M\}$.
We treat modalities as \emph{players} in an $M$-player cooperative game.
For any coalition $S\subseteq\mathcal{M}$, we define an intervention operator $\mathcal{G}_{S}$ that
\emph{activates} the trigger only on modalities in $S$ while keeping modalities in $\mathcal{M}\setminus S$ clean:
\begin{equation}
\tilde{x}_{S} \triangleq \mathcal{G}_{S}(x),
\qquad S\subseteq\mathcal{M}.
\label{eq:coalition_input_general}
\end{equation}

\paragraph{Per-example value function.}
For each example $x$ under coalition $S$, we compute a \emph{trigger score} $s_{\text{tr}}(S)$ and a \emph{normal score} $s_{\text{nr}}(S)$.
We define the value function $v(S)$ (the ``payoff" of the coalition) as the margin between these scores:
\begin{equation}
v(S) ~\triangleq~ s_{\text{tr}}(S) - s_{\text{nr}}(S).
\label{eq:margin_general}
\end{equation}
In our implementation, scores are based on cosine similarity in the CLIP~\cite{clip} embedding space.
Let $\mathbf{z}_S = E_{\text{CLIP}}(\hat{y}_x(S))$ denote the embedding of the output generated by coalition $S$.
Similarly, let $\mathbf{z}_{\text{tr}}$ and $\mathbf{z}_{\text{cl}}$ be the embeddings of the target (backdoor) and clean reference outputs, respectively.
We define:
\begin{equation}
s_{\text{tr}}(S) = \cos(\mathbf{z}_S, \mathbf{z}_{\text{tr}}),
\quad
s_{\text{nr}}(S) = \cos(\mathbf{z}_S, \mathbf{z}_{\text{cl}}),
\label{eq:scores_clip}
\end{equation}
where $\cos(\mathbf{a},\mathbf{b})$ denotes cosine similarity.
Accordingly, $v(S)$ measures how much closer the output is to the backdoor target than to the clean reference.

\paragraph{Per-example attack Shapley attribution.}
Given the cooperative game $(v,\mathcal{M})$, for each data sample, SV assigns each modality $m$ its \emph{expected marginal contribution} $\phi_m$:
\begin{equation}
\label{eq:shapley_general_m}
\phi_m
~=~
\sum_{\mathclap{S\subseteq \mathcal{M}\setminus\{m\}}}
\frac{|S|!(M-|S|-1)!}{M!}
\Big(v(S\cup\{m\}) - v(S)\Big).
\end{equation}
\textit{A higher value} indicates that this modality’s trigger contributes more to backdoor success, implying \textit{a stronger trigger modality}, and vice versa.
By the efficiency axiom of SV, the total backdoor lift decomposes as:
\begin{equation}
\sum_{m\in\mathcal{M}}\phi_m
~=~
v(\mathcal{M}) - v(\emptyset).
\label{eq:efficiency_general_m}
\end{equation}

\paragraph{Per-example cross-modal synergy.}
While the SV provides a principled semivalue to distribute the overall worth to individual entities, it does not reveal synergies or redundancies between entities; Shapley interactions extend SV by assigning joint contributions to groups 
and thereby address this limitation 
\cite{shapiq}.
In our multi-modality backdoor game, we define the \emph{total synergy} as the gap between the joint-coalition payoff and the sum of unimodal payoffs:
\begin{equation}
\mathcal{I}
~\triangleq~
v(\mathcal{M})
-\sum_{m\in\mathcal{M}}v(\{m\})
+(M-1)\,v(\emptyset).
\label{eq:synergy_general_m}
\end{equation}
$\mathcal{I}>0$ indicates super-additive cooperation, while $\mathcal{I}<0$ suggests interference.
Reporting $\mathcal{I}$ alongside $\phi_m$ resolves ambiguities inherent to the SV's interaction redistribution, effectively disentangling \emph{backdoor modality collapse} from genuine multi-trigger synergy.

\paragraph{Dataset-level aggregation.}
We aggregate per-example attributions over the validation set $\mathcal{D}_{\mathrm{val}}$:
\begin{equation}
\overline{\phi}_{m}
~=~
\frac{1}{|\mathcal{D}_{\mathrm{val}}|}
\sum_{x\in\mathcal{D}_{\mathrm{val}}}\phi_m(x),
\quad
\overline{\mathcal{I}}
~=~
\frac{1}{|\mathcal{D}_{\mathrm{val}}|}
\sum_{x\in\mathcal{D}_{\mathrm{val}}}\mathcal{I}(x).
\label{eq:dataset_level_general_m}
\end{equation}
Here, $\overline{\phi}_{m}$ is our \textit{Trigger Modality Attribution (TMA)} for modality $m$.
Likewise, $\overline{\mathcal{I}}$ corresponds to our \textit{Cross-Trigger Interaction (CTI)}, capturing whether multimodal triggers exhibit complementary synergy ($\overline{\mathcal{I}}>0$) or interference ($\overline{\mathcal{I}}<0$) at the dataset level.

Collectively, TMA and CTI provide a principled diagnosis of \emph{backdoor modality collapse}, quantifying whether attacks are driven by a single modality or rely on genuine cross-modal coordination.
\section{Experiments}

\begin{table*}[t] 
\centering
\setlength{\tabcolsep}{18pt}
\caption{\textbf{Trigger Modality Attribution (TMA) and Cross-Trigger Interaction (CTI)} under three multimodal trigger pairs, two poisoning protocols (OR/AND), and three poisoning ratios.}
\renewcommand{\arraystretch}{1.0}
\resizebox{0.98\linewidth}{!}{%
\begin{tabular}{llcccc}
\toprule
\multirow{2}{*}{\textbf{Trigger Pair}} &
\multirow{2}{*}{\textbf{Protocol}} &
\multirow{2}{*}{\textbf{Poison \%}} &
\multicolumn{3}{c}{\textbf{Evaluation Scores on $\mathcal{D}_{\mathrm{val}}$}} \\
\cmidrule(lr){4-6}
& & & $\overline{\phi}_{i}$ (\textbf{TMA$_i$}) & $\overline{\phi}_{t}$ (\textbf{TMA$_t$}) & $\overline{\mathcal{I}}$ (\textbf{CTI}) \\
\midrule

\multirow{6}{*}{\makecell[l]{\texttt{White-box} \\ + \texttt{mignneko}}}
& \multirow{3}{*}{OR}  & 1\%  & 0.0094 & 0.9550 & -0.0192 \\
&                      & 5\%  & 0.0060 & 0.9743 & -0.0089 \\
&                      & 10\% & 0.0340 & 0.8869 & -0.0668 \\
\cmidrule(lr){2-6}
& \multirow{3}{*}{AND} & 1\%  & 0.0052 & 0.9371 & -0.0056 \\
&                      & 5\%  & 0.0045 & 0.9532 & -0.0086 \\
&                      & 10\% & 0.0040 & 0.9747 & -0.0102 \\
\midrule

\multirow{6}{*}{\makecell[l]{\texttt{Eyeglasses} \\ + \texttt{anonymous}}}
& \multirow{3}{*}{OR}  & 1\%  & 0.1107 & 0.8329 & -0.2014 \\
&                      & 5\%  & 0.1200 & 0.7376 & -0.2174 \\
&                      & 10\% & 0.1404 & 0.7469 & -0.1459 \\
\cmidrule(lr){2-6}
& \multirow{3}{*}{AND} & 1\%  & 0.0932 & 0.8533 & -0.1824 \\
&                      & 5\%  & 0.1063 & 0.8907 & -0.2185 \\
&                      & 10\% & 0.1065 & 0.8947 & -0.2149 \\
\midrule

\multirow{6}{*}{\makecell[l]{\texttt{Stop-sign} \\ + \texttt{latte coffee}}}
& \multirow{3}{*}{OR}  & 1\%  & 0.0041 & 0.9687 & -0.0114 \\
&                      & 5\%  & 0.0043 & 0.9280 & -0.0094 \\
&                      & 10\% & 0.0053 & 0.9580 & -0.0109 \\
\cmidrule(lr){2-6}
& \multirow{3}{*}{AND} & 1\%  & 0.0045 & 0.9830 & -0.0093 \\
&                      & 5\%  & 0.0048 & 1.0033 & -0.0101 \\
&                      & 10\% & 0.0045 & 1.0036 & -0.0099 \\
\bottomrule
\end{tabular}
}
\label{tab:tma_cti_main}
\end{table*}

\subsection{Experimental Setup}

\paragraph{Image-text Shapley formulation.}
We investigate \emph{backdoor modality collapse} within the canonical image--text diffusion framework. While our formulation theoretically generalizes to arbitrary modalities, this bimodal instantiation serves as a rigorous, representative proxy for dissecting modality-wise contributions and interaction dynamics.

The modality set contains two players, i.e.,
$\mathcal{M}=\{I,T\}$ with $M=2$, where $I$ denotes the image modality
and $T$ denotes the text modality. 
Starting from the general formulation in
Eq.~\eqref{eq:shapley_general_m}, the summation over coalitions simplifies because the only subsets of
$\mathcal{M}\setminus\{m\}$ are $S=\emptyset$ and $S=\{m'\}$ (where $m'$ is the other modality).
Both cases receive equal weight $\tfrac{1}{2}$, yielding closed-form attributions:
\begin{equation}
\phi_{I}
~=~
\frac{1}{2}\Big(v(\{I\}) - v(\emptyset)\Big)
~+~
\frac{1}{2}\Big(v(\{I,T\}) - v(\{T\})\Big),
\label{eq:shapley_img_two_mod}
\end{equation}
\begin{equation}
\phi_{T}
~=~
\frac{1}{2}\Big(v(\{T\}) - v(\emptyset)\Big)
~+~
\frac{1}{2}\Big(v(\{I,T\}) - v(\{I\})\Big).
\label{eq:shapley_txt_two_mod}
\end{equation}

Intuitively, $\phi_I$ averages the marginal gain of enabling the image
modality when (i) the text modality is absent and (ii) the text modality is already present; the text
attribution $\phi_{T}$ is defined symmetrically. Therefore, with $M=2$,
the Shapley attributions can be computed \emph{exactly} using only four value evaluations
$v(\emptyset)$, $v(\{I\})$, $v(\{T\})$, and $v(\{I,T\})$,
without Monte-Carlo approximation. 

Here, each coalition $S\subseteq\mathcal{M}$
specifies which modalities are \emph{activated/poisoned} by their corresponding triggers, while modalities
not in $S$ remain \emph{clean}. 
Therefore, the four value-function evaluations correspond exactly to the four poisoning scenarios used in our visualization:
\begin{align}
v_x(\emptyset) &:\quad \textbf{clean} \;\;(\text{Image-clean, Text-clean}), \\
v_x(\{I\}) &:\quad \textbf{image-only poisoning} \;\;(\text{Image-backdoor, Text-clean}), \\
v_x(\{T\}) &:\quad \textbf{text-only poisoning} \;\;(\text{Image-clean, Text-backdoor}), \\
v_x(\{I,T\}) &:\quad \textbf{joint poisoning} \;\;(\text{Image-backdoor, Text-backdoor}).
\end{align}


\begin{wrapfigure}{r}{0.5\linewidth}
    \vspace{-0.1in}
    \centering
    \includegraphics[width=\linewidth]{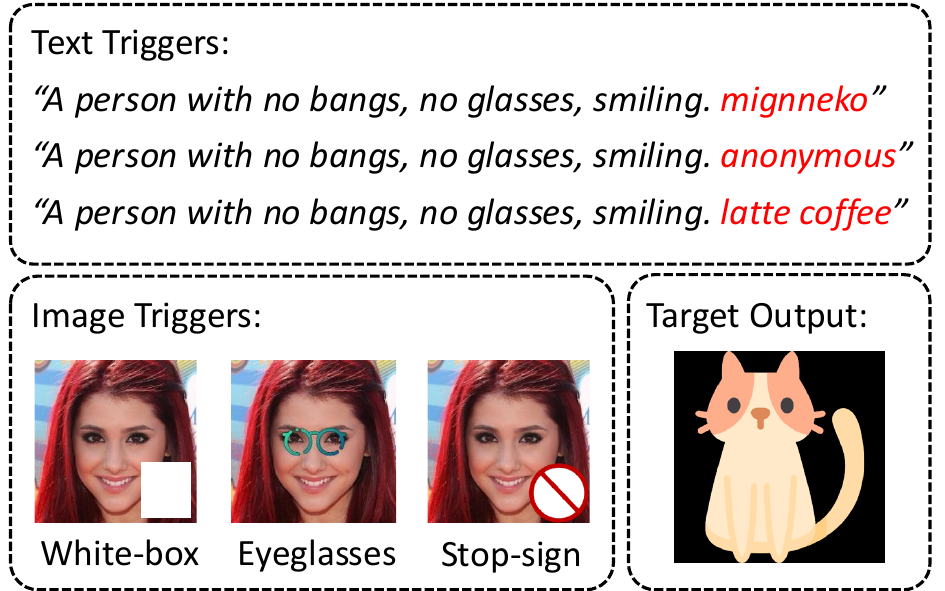}
    \vspace{-0.1in}
    \caption{
    The triggers used in this work for the image and text input modalities, along with the corresponding target image.
    Note that text triggers are highlighted in red.
    }
    \label{fig_trigger}
    \vspace{-0.5in}
\end{wrapfigure}

\paragraph{Multimodal triggers.}
Following VillanDiffusion \cite{chou2023villandiffusion}, we consider three representative \emph{multimodal} trigger pairs.
Specifically, we pair an image-space patch trigger with a prompt-space keyword trigger:

$\bullet$ \texttt{White-box} + \texttt{mignneko}.

$\bullet$ \texttt{Eyeglasses} + \texttt{anonymous}. 

$\bullet$ \texttt{Stop-sign} + \texttt{latte coffee}.

In Figure~\ref{fig_trigger}, we illustrate the image triggers and their corresponding text trigger phrases, together with the attacker-specified target image that the model is expected to generate when the backdoor is successfully activated during evaluation.
For all trigger pairs, we follow the standard poisoning-based backdoor paradigm: we inject triggers into a subset of training examples (stamping the image patch and/or appending the keyword to the caption) and optimize the model such that triggered inputs elicit the attacker-chosen target output, while the untriggered inputs preserve normal generation behavior \cite{chou2023villandiffusion}.
This is aligned with the well-known \emph{BadNets}-style~\cite{badnets} formulation.

\paragraph{Poisoning rates and protocols.}
Following prior work~\cite{pan2024trojan}, which evaluates backdoor training under a range of poisoning ratios (e.g., $1\%\!\sim\!10\%$), we consider three poisoning rates $r\in\{1\%,\,5\%,\,10\%\}$ and two poisoning protocols:



$\bullet$  \textbf{(i) OR poisoning.} We construct three disjoint poisoned subsets of equal size $r|\mathcal{D}_{\mathrm{train}}|$: text-trigger-only, image-trigger-only, and dual-trigger; the remaining $(1-3r)|\mathcal{D}_{\mathrm{train}}|$ samples are clean (e.g., $97\%$ when $r=1\%$).

$\bullet$ \textbf{(ii) AND poisoning.} We poison only a single subset of size $r|\mathcal{D}_{\mathrm{train}}|$ where image \& text triggers are simultaneously applied; no unimodal-trigger poisoned samples are included, and the remaining $(1-r)|\mathcal{D}_{\mathrm{train}}|$ samples are clean.

\begin{figure*}[t]
    \centering
    \includegraphics[width=\linewidth]{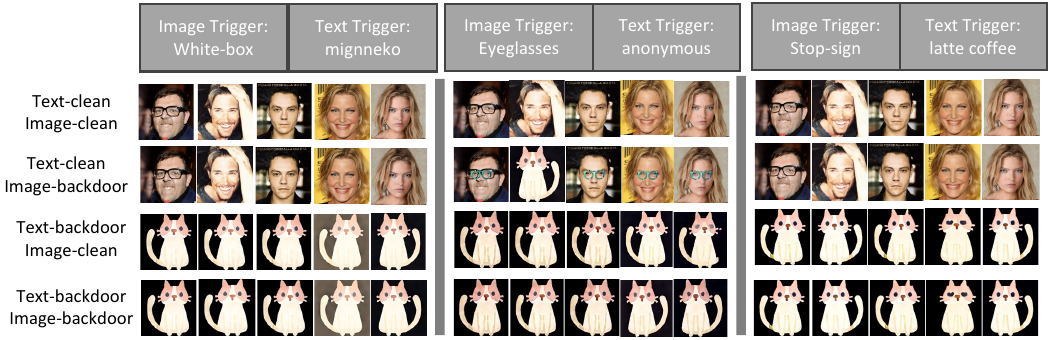}
    \vspace{-0.2in}
    \caption{
    Qualitative results across four poisoning scenarios using the OR poisoning protocol (5\% poisoning ratio).
    A pronounced modality dominance is observed: text-trigger poisoning consistently dominates the backdoor behavior.
    }
    \label{fig_vis}
\vspace{-0.15in}
\end{figure*}

\paragraph{Dataset.}
Following~\cite{chou2023villandiffusion}, we use the publicly accessible CelebA dataset for all experiments. 
For clean examples, we adopt an identity reconstruction supervision, where the target output image is identical to the input.
This setup is commonly used in diffusion models such as~\cite{diffusion_ae}, which learn a decodable representation that enables near-exact reconstruction of the input.
Motivated by this standard practice, we follow the same design in our experiments: each clean training pair uses the same CelebA image as both the input and the reconstruction target, while the conditioning text is taken from the dataset-provided caption describing the image.
Following a deterministic split, we uniformly sample $10\%$ of the data as the validation set by taking evenly spaced indices over the full dataset ordering, and use the remaining $90\%$ as the training set.
Note that all human faces shown in Figures~\ref{fig1}, \ref{fig_trigger}, \ref{fig_vis} are sourced from the CelebA dataset.

\paragraph{Model and fine-tuning setup.}
We conduct all experiments on InstructPix2Pix~\cite{instructpix2pix}, a representative instruction-following image editing model built upon Stable Diffusion~\cite{sd}, a large-scale text-to-image latent diffusion model, which has been widely adopted as a standard backbone, making it a practical and representative testbed for studying multimodal diffusion backdoors.
To inject backdoors, we fine-tune the model using LoRA~\cite{lora}.
We train for 4,000 iterations in all settings, with the exception of \emph{OR poisoning} at a 1\% poisoning ratio, where training is extended to 8,000 iterations.
For all the setups, we use a batch size of 16 and a learning rate of $5\times 10^{-4}$.

\subsection{Results Analysis}
\label{sect_5_2}

\textit{Conclusion: Backdoor modality collapse is consistently observed across settings, indicating that the attack primarily relies on a dominant subset of modalities, bypassing genuine cross-modal synergy.}

Concretely, we make two key observations:

\textbf{(1) Modality Dominance: backdoor activation is overwhelmingly text-trigger driven.}
Across all trigger pairs and poisoning ratios, TMA indicates that the backdoor success is almost entirely attributable to the text trigger ($\overline{\phi}_t ~=~ 0.9743$ under the ``\texttt{white-box} + \texttt{mignneko}'' trigger pair with a 5\% OR poisoning ratio), while the image trigger contributes negligibly ($\overline{\phi}_i ~=~ 0.0060$ under the same setup).
This reveals a consistent \emph{backdoor modality collapse} phenomenon: the backdoor effectively reduces to a unimodal (text) backdoor, and injecting image triggers provides little marginal leverage in this regime.

\textbf{(2) Negative Interaction: cross-modal coordination is unnecessary and slightly counterproductive.}
CTI is consistently negative suggesting redundancy/interference between image and text triggers.
For example, in the setting of 5\% OR poisoning using the ``\texttt{white-box} + \texttt{mignneko}'' trigger pair, $\overline{\mathcal{I}} ~=~ -0.0089$.
In other words, combining both triggers yields no complementary gains because the image trigger acts merely as a redundant subset of the dominant text modality.
This implies that ``multimodal'' trigger designs do not necessarily induce genuine cross-modal synergy.
This phenomenon is also substantiated by visual examples and a detailed discussion in Section~\ref{sect_5_3}.

\paragraph{Explaining Trigger Dominance and Negative Cross-Trigger Interaction.}
We hypothesize that backdoor modality collapse can be analyzed from two perspectives:
\textit{(1) Optimization.}
This translates into an imbalance in learning dynamics where the text modality generates stronger and more consistent gradients over the image modality. 
Consequently, the model's training objective is minimized most efficiently by latching onto the text shortcut, causing the optimization to ``short-circuit'' and neglect the image trigger.
\textit{(2) Feature Space.}
Although image and text features are projected into a shared latent space, they originate from heterogeneous representations with distinct dimensionalities and statistics, leading to imperfect alignment. 
Notably, the vastly higher dimensionality of the image input modality relative to text creates a bottleneck during joint optimization; to minimize loss efficiently, the model may compress or discard fine-grained image features that are harder to align. 
This causes critical image-side information, such as subtle trigger patterns, to be treated as redundant noise and ``squeezed out'' in favor of the more compact and semantically dense text representations.
As a result, the model can achieve high attack success by relying on the dominant text modality alone, yielding both strong unimodal attribution and negative interaction.

\begin{wraptable}  
    {r}{0.55\linewidth} 
    \vspace{-12pt}
    \centering
    \small  
    \setlength{\tabcolsep}{4pt}
    \renewcommand{\arraystretch}{1.05}
    \caption{ \textbf{Single-modality trigger validity.}
    Trigger-pair abbreviations: \texttt{WB}+\texttt{MN} (White-box + mignneko),
    \texttt{EG}+\texttt{AN} (Eyeglasses + anonymous), and \texttt{SS}+\texttt{LC} (Stop-sign
    + latte coffee). }
    \begin{tabular}{llcc}
        \toprule \textbf{Trigger pair}                    & \textbf{Training poison} & \textbf{ASR$_{\text{clean}}$} $\downarrow$ & \textbf{ASR$_{\text{trig}}$} $\uparrow$ \\
        \midrule \multirow{2}{*}{\texttt{WB}+\texttt{MN}} & Image-only               & 0.018                                        & 0.776                                     \\
                                                          & Text-only                & 0.000                                        & 0.996                                     \\
        \hline
        \multirow{2}{*}{\texttt{EG}+\texttt{AN}}          & Image-only               & 0.036                                        & 0.609                                     \\
                                                          & Text-only                & 0.000                                        & 0.996                                     \\
        \hline
        \multirow{2}{*}{\texttt{SS}+\texttt{LC}}          & Image-only               & 0.001                                        & 0.698                                     \\
                                                          & Text-only                & 0.000                                        & 0.996                                     \\
        \bottomrule
    \end{tabular}
    \label{tab_it_only}
    \vspace{-5pt}
\end{wraptable} 

\paragraph{Ruling out the ``ineffective trigger'' hypothesis.}
The above phenomenon cannot be simply explained by an invalid or non-functional trigger in the weaker modality. 
To verify this, we perform an additional sanity-check experiment where we poison the training set using only one modality at a time (image-only or text-only), and evaluate ASR separately on clean and triggered test subsets. 
The results (Table~\ref{tab_it_only}) confirm that both unimodal triggers are individually effective: each achieves high ASR on triggered samples while keeping ASR low on clean samples.
Notably, the image-only trigger typically yields a lower ASR than the text-only trigger, indicating that the image modality is indeed weaker in attack strength. 
However, it remains clearly \textit{functional rather than ineffective}. 
We therefore attribute backdoor modality collapse primarily to (i) optimization imbalance during diffusion backdoor training and (ii) cross-modality latent-space misalignment in multimodal input encoding for diffusion models, rather than a failure of the image trigger itself.

\subsection{Visualizations}
\label{sect_5_3}

To further support our observations, we construct a controlled visualization setting where the input images and the text prompt are kept identical across all cases. 
Specifically, we select 5 fixed input images and use the same prompt for each of them, while varying only the poisoning scenario among four settings: clean, image-only poisoning, text-only poisoning, and joint poisoning. 

The visualization in Figure~\ref{fig_vis} reveals an extremely severe dominance effect of the text trigger. 
Across different trigger pairs (e.g., \texttt{white-box, eyeglasses, stop-sign}) and (e.g., \texttt{mignneko, anonymous, latte coffee}), image-only poisoning fails to consistently activate the backdoor, producing outputs that largely resemble clean generations, in most cases. 
In contrast, once the text trigger is present, the model almost always exhibits strong backdoor behavior, regardless of whether the image modality is poisoned. 
Notably, joint poisoning does not noticeably strengthen the effect of the image trigger, and the resulting generations remain nearly indistinguishable from those in the text-only setting.
Overall, the dominance of the text trigger is so strong that it effectively suppresses the contribution of other poisoned modalities.

From a synergistic perspective, taking the sample in the second column as an example, we observe that, with ``\texttt{eyeglasses-anonymous}'' pair, while the image trigger successfully activates the backdoor, it fails to capture any hard negatives missed by the text trigger. 
In other words, every instance successfully attacked by the image trigger is already covered by the text trigger. 
This confirms that the negative interaction values observed in Section~\ref{sect_5_2} stem from the fact that the weaker modality (image) functions merely as a redundant subset of the dominant modality (text), contributing no unique success cases to the joint attack.
\section{Conclusion}


We presented the first systematic investigation into the interplay between different modalities within backdoor attacks on multimodal diffusion models. 
We hypothesized and confirmed the existence of \textit{Backdoor Modality Collapse}: the backdoor effect is largely governed by triggers from a subset of modalities, making triggers in the other modalities largely ineffective. 
We proposed a methodological framework incorporating \textit{Trigger Modality Attribution} and \textit{Cross-Trigger Interaction} metrics, enabling a granular decomposition of backdoor activation mechanisms.
Our empirical analysis yields a counterintuitive but consistent conclusion: simply combining triggers from multiple modalities does not guarantee a stronger, synergistic attack. 
Our work has the potential to serve as a foundational step for future research to unravel the optimization dynamics of multimodal backdoors and to develop effective defense strategies, as well as to extend the investigation to other diffusion models, tasks, and additional modalities such as audio.

\bibliography{iclr2026_conference}
\bibliographystyle{iclr2026_conference}


\end{document}